\documentclass[runningheads]{llncs}
\usepackage{array, makecell}
\usepackage{multirow}
\usepackage{textcomp}
\usepackage{listings}
\usepackage{cleveref}
\usepackage{bold-extra}
\usepackage{rotating}
\usepackage{makecell}
\usepackage{xcolor}

\makeatletter
\newcommand{\thickhline}{%
    \noalign {\ifnum 0=`}\fi \hrule height 1pt
    \futurelet \reserved@a \@xhline
}
\newcolumntype{"}{@{\hskip\tabcolsep\vrule width 1pt\hskip\tabcolsep}}
\makeatother

\begin{document}
\title{R2RML and RML Comparison for\\
RDF Generation, their Rules Validation\\
and Inconsistency Resolution
}
\titlerunning{R2RML and RML Comparison for
	RDF Generation and their Rules Validation}
%
\author{Anastasia Dimou\orcidID{0000-0003-2138-7972}}
\authorrunning{A. Dimou et al.}
%
\institute{Ghent University -- imec -- IDLab, Belgium}
\maketitle              
\begin{abstract}
In this paper,
an overview of the state of the art
on knowledge graph generation is provided,
with focus on the two prevalent mapping languages:
the W3C recommended R2RML and 
its generalisation RML.
We look into details on their differences
and explain how knowledge graphs, in the form of RDF graphs,
can be generated with each one of the two mapping languages.
Then we assess
if the vocabulary terms were properly applied to the data
and no violations occurred on their use,
either using R2RML or RML to generate the desired knowledge graph.


\end{abstract}

\section{Introduction}

Knowledge graphs are often generated using
rules that apply vocabulary terms to certain data
from different data sources.
This occurs because
most data is still not available in the form of knowledge graphs.
Data may have \textit{different structures}
(e.g., tabular or hierarchical),
appear in \textit{heterogeneous formats}
(e.g., CSV, XML or JSON)
and are accessed via \textit{heterogeneous interfaces}
(e.g., database interfaces or Web APIs)~\cite{Dimou_2015_LogicalSource}.
Therefore, different approaches were proposed
to generate knowledge graphs
from existing (semi-)structured data.

These different approaches follow different directions,
from \textit{custom implementations} \cite{Calvanese_2011,Haase2019Hybrid}
to \textit{format-specific} \cite{Langegger_2009_XLWrap,Lange_2011_Krextor}
or \textit{direct mappings} \cite{Arenas2012,scharffe:hal-00768424},
but approaches that \textit{detach the rules definition
from their execution} prevailed.
Detaching the rules renders the rules interoperable between implementations,
whilst the systems that process those rules are use-case independent.
This improves the knowledge graph generation
which becomes
more interoperable, reusable, and maintainable.
\textit{Mapping languages} were proposed
to define such rules to generate knowledge graphs
which range from
\textit{dedicated languages}, such as R2RML, RML or xR2RML~\cite{dimou_ldow_2014,michel2015translation}
to \textit{repurposed languages}, such as SPARQL-Generate~\cite{lefrancois_eswc_2017}.

Even though many dedicated mapping languages were proposed,
no thorough comparison between those languages
was performed so far.
Only De Meester et al.~\cite{DeMeester2019Mapping}
presented an initial set of comparative characteristics based on requirements posed by reference works.
These characteristics are both
\textit{functional}
(easy to use \cite{Heyvaert2018Declarative,lefrancois_eswc_2017},
follow Semantic Web standards~\cite{lefrancois_eswc_2017}, 
fully cover the generation process \cite{ademar2016FunUL})
and \textit{non-functional}
(be extensible \cite{dimou_ldow_2014,ademar2016FunUL,lefrancois_eswc_2017} and
support general mapping functionalities \cite{ademar2016FunUL},
nested hierarchies \cite{michel2015translation}
and lists \cite{michel2015translation}).
De Meester et al.~\cite{DeMeester2019Mapping}
observe that multitude of mapping languages
allows to support more use cases and
conclude that
effort should be consolidated on missing features,
instead of re-developing existing functionalities.

In this paper,
we compare the two most prevalent mapping languages
with respect to their syntax:
the W3C recommended R2RML~\cite{R2RML} and
its generalisation RML~\cite{dimou_ldow_2014}.
We compare what can be done and how
in each mapping language.
We aim to support data holders to answer
the following research question:



\begin{quote}
    \textit{Which mapping language between R2RML and RML should one use?}
\end{quote} 

The answer to this research question is explored in \cref{sec:MappingLanguages}.

\vspace{1em}



However, while knowledge graphs are generated
by consistently applying terms of certain vocabularies,
the rules that define
how the vocabularies terms are applied should respect
the restrictions imposed by the vocabularies definitions.
However, consistently annotating
the existing data sources with vocabulary terms
is not always straightforward.
Inconsistencies may be introduced
when the rules are defined \cite{dimou2015assessing}
and resolutions are required \cite{Heyvaert2019Rule},
applied to either the rules or the vocabulary terms definitions~\cite{Heyvaert2019Rule}.
We aim to support data holders
to answer the following research question:

\begin{quote}
    \textit{How can one validate that the vocabulary terms
    were properly applied to the data using R2RML or RML?}
\end{quote}

The answer to this research question is explored in \cref{sec:Validation}.

The remaining of this paper is structured as follows:
In \cref{sec:SotA}, we outline the state of the art
on knowledge graph generation using mapping languages,
and on knowledge graph validation
and inconsistency resolution.
Then we focus on R2RML and RML mapping languages
and we compare them in \cref{sec:MappingLanguages}.
Last, we explain how rules expressed with these mapping languages
can be validated and how inconsistencies can be resolved.
The paper concludes with a discussion on
remaining challenges, open issues and future directions.


\section{State of the Art}
\label{sec:SotA}

This paper focuses on (i)~\textit{defining} rules
to annotate data with vocabulary terms
using mapping languages on the one hand,
and (ii)~\textit{validating} rules
that annotate data  on the other hand.
In \cref{subsec:mappingLanguages},
the state of the art related to
mapping languages is covered,
while in \cref{subsec:validationSotA},
the state of the art on
validation and resolution approaches
for mapping languages is covered.

\subsection{Mapping Languages}
\label{subsec:mappingLanguages}

Different approaches were proposed so far
for generating knowledge graphs
from (semi-)structured data sources.
These approaches range from \textit{custom implementations} \cite{Calvanese_2011,Makela_2012},
which appeared mostly in the first years
but remain prevalent till nowadays \cite{Catherine_2019_WeatherDP,Haase2019Hybrid},
to more \textit{generic approaches}.
Such generic approaches
originally focused on data with \textit{specific formats},
namely dedicated approaches for,
e.g.,~relational databases~\cite{D2RQ}, or XML~\cite{Lange_2011_Krextor}.
Among the format-specific approaches,
solutions for knowledge graph generation from relational databases
were the most mature ones and lead to two main directions
which also became W3C recommendations:
(i) \textit{direct mapping}~\cite{Arenas2012}
and (ii) \textit{detached rules}~(R2RML)~\cite{R2RML}.

In the former case, the \textit{direct mapping}
defines a simple transformation,
providing a basis for
defining more complex transformations afterwards.
However,
\textit{direct mapping} requires defining rules later,
for instance using SPARQL queries~\cite{sparql_spec},
e.g., Datalift \cite{scharffe:hal-00768424},
to replace the original predefined annotations with custom ones.
In the latter case,
the rules definition is detached from their execution.
Mapping languages were proposed
to define such detached rules,
such as D2RQ~\cite{D2RQ} that lead to
the W3C recommended R2RML~\cite{R2RML}.

However, format-specific approaches,
like the ones for relational databases,
require data holders to learn and maintain
different tools for each data format~\cite{dimou_ldow_2014}.
Therefore, 
different solutions were proposed for
heterogeneous data sources.
The prevalent directions for knowledge graph generation
from relational databases, were followed.
However, the Direct Mapping cannot be applied for
heterogeneous data sources,
because each format requires
its own Direct Mapping implementation,
as it occurs in the case of Datalift~\cite{scharffe:hal-00768424}.

Therefore,
most solutions for knowledge graph generation
from heterogeneous data sources were focused on detached rules.
On the one hand,
\textit{dedicated mapping languages} were proposed, e.g., RML~\cite{dimou_ldow_2014}
that generalises R2RML or
xR2RML~\cite{michel2015translation,michel2016xr2rml}
that extends both R2RML and RML.
On the other hand,
\textit{repurposed mapping languages}
were proposed that extend
existing languages for other tasks,
e.g., SPARQL-Generate \cite{lefrancois_eswc_2017}
that repurposes SPARQL~\cite{sparql_spec}.
Nowadays,
the most prevalent dedicated mapping languages are
defined as extensions of R2RML.
RML was the first language to generalise R2RML,
but there are more alternative approaches and extensions
beyond the originally proposed language.


\subsection{Validation and inconsistency resolution}
\label{subsec:validationSotA}

There exists a number of methods to validate knowledge graphs,
and detect and resolve inconsistencies.
The validation might be applied on 
either the knowledge graph
(\textit{graph-driven validation methods})
or the rules (\textit{rule-driven validation methods})
that define how the knowledge graph is generated.

On the one hand,
there are validation methods
applied directly to the knowledge graph.
Different approaches were proposed
to tackle various aspects of Linked Data quality.
These approaches can be classified as \cite{DeMeester2019RDF}:
(i) \textit{manual} (e.g., \cite{Acosta2013,Bizer2009c}),
(ii) \textit{hard-coded} \cite{Hogan:2012:ESL:2263498.2264570}
(iii)~\textit{logic-based} (\textit{integrity constraints}, e.g.,~\cite{Patel-Schneider2014Using}),
(iv)~\textit{query-based}, e.g., \cite{farid2016clams,Kontokostas2014} and
(v)~\textit{rule-based}, e.g.,\cite{DeMeester2019RDF}.
These methods rely on the complete knowledge graph
and can potentially identify every inconsistency.
However,
validating the complete knowledge graph
is not always possible,
especially in time-constrained situations~\cite{dimou2015assessing}.
On the other hand,
there are validation methods applied to the rules
that generate knowledge graphs~\cite{dimou2015assessing}.
Such methods result in faster execution times,
but not all inconsistencies can be identified,
as some of them depend on
the actual data values in the graph,
e.g.,  (qualified) cardinality,
(inverse) functionality,
(a)symmetricity and irreflexivity.

Methods were proposed to
detect and resolve inconsistencies.
 Possible root
causes for these inconsistencies include \cite{Heyvaert2019Rule}:
(i) \textit{raw data}
that contain inconsistencies~\cite{Lehmann2015DBpediaA};
(ii) \textit{rules} that introduce new inconsistencies
by, for example,
not using the suitable ontology terms~\cite{dimou2015assessing,paulheim_eswc_2017};
and
(iii) \textit{vocabulary definitions}
that do not model the domain as desired~\cite{paulheim_eswc_2017}.

\section{Mapping Languages: R2RML and RML}
\label{sec:MappingLanguages}

In this chapter,
we focus on knowledge graph generation
with \textit{dedicated mapping languages}.
The most noted mapping language
is the W3C recommendation R2RML~\cite{R2RML},
and its most popular generalisation RML~\cite{dimou_ldow_2014}
for heterogeneous data sources,
while xR2RML~\cite{michel2016xr2rml}
is their common derivative.
RML may be considered broadly adopted
because it is the R2RML extension
that has the most alternative implementations
according to its implementation report\footnote{RML Implementation Report, \url{https://rml.io/implementation-report}}:
RMLMapper\footnote{RMLMapper, \url{https://github.com/RMLio/rmlmapper-java}},
CARML\footnote{CARML, \url{https://github.com/carml/carml}},
RMLStreamer\footnote{RMLStreamer, \url{https://github.com/RMLio/RMLStreamer}},
RocketML\footnote{RocketML, \url{https://github.com/semantifyit/RocketRML}},
SDM-RDFizer\footnote{SDM-RDFizer, \url{https://github.com/SDM-TIB/SDM-RDFizer}}. 

\paragraph{R2RML}
The Relational to RDF Mapping Language (R2RML)~\cite{R2RML}
is the W3C recommendation
to express customized mapping rules from data in relational databases
to generate knowledge graphs represented
using the Resource Description Framework (RDF)~\cite{RDF}.
R2RML considers a target semantic schema
which is a combination of one or more vocabularies.
The R2RML vocabulary namespace is \texttt{http://www.w3.org/ns/r2rml\#} 
and the preferred prefix is \texttt{r2rml}.

\paragraph{RML}

The RDF Mapping Language (RML)~\cite{dimou_ldow_2014}
also expresses customized mapping rules,
but from data in
heterogeneous structures, formats and access interfaces.
RML is a superset of R2RML, 
but always remains backwards compatible with R2RML. 
RML keeps the mapping rules as in R2RML 
but excludes its database-specific references from the core model. 
This way,
the mapping rules can be defined
in a combined and uniform way, 
while the input data becomes
a set of (one or more) input data sources.
The RML vocabulary namespace is \texttt{http://semweb.mmlab.be/ns/rml\#} 
and the preferred prefix is \texttt{rml}.

\begin{table}[ht]
	\centering
	\caption{R2RML and RML comparison}
	\label{tab:prefixes}
	\begin{tabular}{r|l|l}
	\textbf{language}
	& \textbf{R2RML}
	& \textbf{RML} \\ \hline
	\textbf{prefix}
	& \texttt{rr}
    & \texttt{rml} \\ \hline
    \textbf{URI} 
	& \texttt{http://www.w3.org/ns/r2rml\#} 
	& \texttt{http://semweb.mmlab.be/ns/rml\#}  \\ \hline
	\makecell[r]{\textbf{relational}\\\textbf{DBs}}
	& \makecell[l]{multiples tables \\ one DB}
	& \makecell[l]{multiple tables \\ multiple DBs} \\ \hline
	\textbf{access interfaces}
	& only ODBC
	& multiple\\ \hline
	\makecell[r]{\textbf{other data}\\\textbf{structures}}
	& --
	& \makecell[l]{tabular (e.g., CSV, TSV, XLS) \\
	hierarchical (e.g., XML, JSON)\\
	pair-valued (e.g., wikitext) \\
	graphs (e.g., RDF), etc.}\\ \hline
	\textbf{integration}
	& \makecell[l]{materialisation\\virtualisation}
	& \makecell[l]{materialisation\\virtualisation}\\ \hline
	\makecell[r]{\textbf{data}\\\textbf{transformation}}
	& pre-processing
	& \makecell[l]{pre-processing\\inline processing}\\ \hline
	\end{tabular}
\end{table}

\vspace{1em}

In both R2RML and RML,
an RDF dataset is generated
based on one or more \textit{Triples Maps}
(\texttt{rr:TriplesMap}).
A \textit{Triples Map} defines
how a set of RDF triples
referring to the same entity,
i.e. same subject, are generated.
Each \textit{Triples Maps}
defines 
which data is considered (\cref{subsec:dataSource}),
what the \textit{subject} of the RDF triples is
(\textit{Subject Map}), and
which predicates (\textit{Predicate Map})
and objects (\textit{Object Map})
will be generated to form the different RDF triples
(\cref{subsec:predicateObject}).

The following subsections explain
the differences between R2RML and RML
with respect to
(i) the data sources they support (\cref{subsec:dataSource}),
(ii) the RDF terms generation (\cref{subsec:TermGeneration}), and 
(iii) the RDF triples generation
(\cref{subsec:tripleGeneration}).

\subsection{Data Source}
\label{subsec:dataSource}

R2RML and RML support \underline{different data sources}
to generate the RDF datasets.

\paragraph{R2RML}
A \textit{Triples Map} refers to a \textit{logical table}
retrieved from a certain database.
A logical table can be
a base table,
a view, or
a valid SQL query, called an \textit{R2RML view}
because it emulates a SQL view
without modifying the database.

\paragraph{RML}
Data can originally 
(i)~reside on \textbf{diverse locations},
e.g., in files or in a database
at the local network, or can be published on the Web;
(ii)~be accessed using \textbf{different interfaces},
e.g.~raw files, database connectivity for
databases, different interfaces from the Web, such
as Web APIs; and 
(iii)~have \textbf{heterogeneous structures and formats},
e.g.~tabular, such as databases or CSV files,
hierarchical, such as XML or JSON format,
semi-structured, such as HTML \cite{Dimou_2015_LogicalSource}.

\begin{table}[ht]
	\centering
	\caption{Results of female pole vault for 2019 world championship}
	\label{tab:femalePoleVaulters}
	\begin{tabular}{l|l|l|l|l|l}
	\textbf{rank}
	& \textbf{name}
	& \textbf{surname}
	& \textbf{nationality}
	& \textbf{mark}
	& \textbf{notes} \\ \hline
	1
	& Anzhelika 
	& Sidorova
	& Russia
	& 4.95
	& WL,PB \\ \hline
	2
	& Sandi
	& Morris
	& United States (USA)
	& 4.90
	& SB \\ \hline
	3
	& Katerina
	& Stefanidi
	& Greece
	& 4.85
	& SB \\ \hline
	4
	& Holly 
	& Bradshaw
	& Great Britain
	& 4.80
	& \\ \hline
	
	\end{tabular}
\end{table}


The main difference between R2RML and RML
is the type of data they support.
R2RML supports \underline{homogeneous data}, i.e. relational databases,
whereas RML supports \underline{heterogeneous data}.
In more details:

\paragraph{R2RML} 
Each \textit{Triples Map}
(\cref{lst:r2rml_logicalTable}, \cref{r2rml_triples:tm})
refers to \underline{exactly one \textit{Logical Table}}
(\texttt{rr:LogicalTable}),
specified by its \textit{table name}
(\texttt{rr:tableName}).
A \textit{Logical Table}
is either a SQL base table or view,
or an R2RML view.
The SQL query result
is used to generate the RDF triples.
All mapping rules refer to only one database.

\begin{lstlisting}[label={lst:r2rml_logicalTable},
caption=A Triples Map refers to a certain Logical Table specified by its name]
§\label{r2rml_triples:tm}§<#PoleVaulters>        rr:logicalTable <#PoleVaultersDBtable> ; 
<#PoleVaultersDBtable> rr:tableName "poleVaulters" .

\end{lstlisting}

\paragraph{RML}
A \textit{Logical Source} (\texttt{rml:LogicalSource}) extends R2RML’s \textit{Logical Table} and
defines the data source
to be used for the knowledge graph.
A broader reference to any input source is considered in RML.
Thus, \underline{exactly one \textit{Logical Source}} is specified (\texttt{rml:source})
for each \textit{Triples Map} to indicate the input.
For instance,
instead of having a \textit{Logical Table} \texttt{<\#PoleVaultersDBtable>} (\cref{lst:r2rml_logicalTable}, \cref{r2rml_triples:tm}),
we have a \textit{Logical Source}
\texttt{<\#CountriesXML>} (\cref{lst:rml_logicalSource}, \cref{rml_triples:tm}):

\begin{lstlisting}[label={lst:rml_logicalSource},
caption=A Triples Map refers to a Logical Source whose data is in XML format]
§\label{rml_triples:tm}§<#Countries>    rml:logicalSource <#CountriesXML> . 
<#CountriesXML> rml:source <http://rml.io/data/semWebSer/countries.xml> .
\end{lstlisting}

Such a \textit{Logical Source}
may refer to a file that contains the countries
with the pole vaulters nationality
such as the one that follows (\cref{lst:countries}).

\begin{lstlisting}[label={lst:countries},
caption=An XML file with countries details]
<?xml version="1.0" encoding="UTF-8"?>
<countries>
  <country country_language="en-uk">Great Britain</country>
  <country country_language="el">Greece</country>
  <country country_language="ru">Russia</country>
  <country country_language="en-us">United States</country>
</countries>

\end{lstlisting}

\subsubsection{Data Iteration}
Mapping languages are suitable for
knowledge graphs generation
because the same rules
are applied to multiple data chunks
that follow the same structure pattern.
While R2RML is focused on tabular structure
whose iteration pattern is predefined,
the iteration pattern in other data structures
is not be predefined, causing yet another difference
between R2RML and RML.

\paragraph{R2RML}
An \underline{implied per-row iteration} is predefined.

\paragraph{RML}
The iteration pattern
can not always be implicitly assumed,
but it needs to be \underline{explicitly defined}.
The \textit{iterator} (\texttt{rml:iterator}) (\cref{lst:rml_iter}, \cref{rml_triples:iter})
determines the iteration pattern over the data source
and specifies the data to be considered
during each iteration.
The \textit{iterator} is not required to be explicitly mentioned
in the case of tabular data sources,
as the default per-row iteration is implied.

\begin{lstlisting}[label={lst:rml_iter},
caption=A Logical Source specifies its iterator]
§\label{rml_triples:iter}§<#CountriesXML>   rml:iterator "/countries/country" .
\end{lstlisting}

\subsection{RDF term generation}
\label{subsec:TermGeneration}

To generate the RDF triples of an RDF dataset,
the RDF terms that define
the subject, predicate, object and named graph
need to be generated.

\textit{Term Maps} (\texttt{rr:TermMap}) define
how RDF terms (IRI, blank node, literal) are generated.
They are \textit{constant-}, \textit{template-},
or \textit{column-} (R2RML)
or \textit{reference-} (RML) \textit{valued}.
The subject, predicate and object of RDF triples
are RDF terms.

A \textit{constant-valued Term Map} (\texttt{rr:constant}, \cref{lst:r2rml_rml_termMap}, \cref{r2rml:cv})
always generates the same RDF term
which is by default an IRI.

A \textit{template-valued} Term Map (\texttt{rr:template}, \cref{lst:r2rml_rml_termMap}, \cref{r2rml:template})
is a valid string template that contains references
and generates an IRI by default.

\begin{lstlisting}[label={lst:r2rml_rml_termMap},
caption=RDF terms are generated with different types of Term Maps]
§\label{r2rml:cv}§<#RDFtermConstant>       rr:constant          ex:score .
§\label{r2rml:template}§<#RDFtermTemplate>       rr:template         "http:://ex.com/person/{name}_{surname}" .
\end{lstlisting}

\subsubsection{Reference to data fragments}

R2RML and RML refer differently 
to the data fragments of the data sources
based on which the RDF terms are generated.

\paragraph{R2RML}
A \textit{\underline{column-valued} term map}
(\texttt{rr:column}, \cref{lst:r2rml_termMap}, \cref{r2rml:column})
generates a literal by default
that is a column in a given \textit{Logical Table}’s row.

\begin{lstlisting}[label={lst:r2rml_termMap},
caption=RDF terms are generated with different types of Term Maps]
§\label{r2rml:column}§<#RDFtermColumn>       rr:column "name" .

\end{lstlisting}

\paragraph{RML}
As RML covers heterogeneous data,
different references to the data sources apply.
A \textit{logical reference} (\texttt{rml:reference})
is a valid reference to the data source
according to the specified reference formulation.
The \textit{Reference Formulation} (\texttt{rml:referenceFormulation}) (\cref{lst:rml_refForm}, \cref{rml_triples:refForm}), indicates the formulation
(for instance, a standard or a query language) used to refer to its data. 
A \textit{\underline{reference-valued} term map}
generates a literal by default. 

\begin{lstlisting}[label={lst:rml_refForm},
caption=A Logical Source specifies its Reference Formulation]
§\label{rml_triples:refForm}§<#CountriesXML>      rml:referenceFormulation ql:XPath .
§\label{rml:reference}§<#RDFtermReference>  rml:reference "name" .
\end{lstlisting}

\subsubsection{Language}

The language of an RDF term
which is literal can be defined.
R2RML and RML define differently the language,
RML extends the R2RML options.

\paragraph{R2RML}
The language (\texttt{rr:language}) of a literal
may be optionally defined.
R2RML allows \underline{only constant values} for the language.

\begin{lstlisting}[label={lst:r2rml_language},
caption=rr:language to define the literal's language]
<#CountryName>  rr:column "country" ;
§\label{rr:language}§                rr:language "en-us" .

\end{lstlisting}

\paragraph{RML}
RML has a dedicated \textit{Term Map}
defining the language,
the \textit{Language Map} (\texttt{rml:LanguageMap},
\cref{lst:rml_language}, \cref{rml:languageMap}), 
which extends R2RML's language tag (\texttt{rr:language}).
The \textit{Language Map} allows
not only constant values for defining the language
but \underline{also references to the data}.
\texttt{rr:language} is considered then
an abbreviation for the \texttt{rml:languageMap},
if the value is constant.

\begin{lstlisting}[label={lst:rml_language},
caption=rml:languageMap to define the literal's language]
§\label{r2rml:pm_abb}§<#CountryName>  rml:reference "country_name" ;
§\label{rml:languageMap}§                rml:languageMap [ rml:reference "@country_language"] ] .

\end{lstlisting}

\subsubsection{datatype}
The datatype (\texttt{rr:datatype}) of an RDF term
which is literal can be defined as well.
Both R2RML and RML
define the \textit{datatype} \underline{in the same way}.

\begin{lstlisting}[label={lst:r2rml_datatype},
caption=rr:datatype to define the literal's datatype]
<#Mark>         rr:column "mark" ;
§\label{rr:datatype}§                rr:datatype xsd:positiveInteger .
\end{lstlisting}

\subsubsection{Term type}
If the default termtype is desired to be changed,
the term type (\texttt{rr:termType})
is explicitly defined (\texttt{rr:IRI}, \texttt{rr:Literal}, \texttt{rr:BlankNode}).

The \textit{term type} (\texttt{rr:termType})
needs to be explicitly mentioned
(\cref{lst:r2rml_termType}, \cref{rr:termType})
to change the default term type, i.e.
\texttt{IRI} for \textit{template}
and \textit{constant-}valued term maps
and \texttt{Literal}
for \textit{column-} or \textit{reference-}valued term maps.
\Cref{tab:termTypeCombinations} summarizes
all valid combinations of \textit{Term Maps}
and \textit{Term Types} and the type of \textit{RDF term}
they generate in each case for both R2RML and RML.

\begin{lstlisting}[label={lst:r2rml_termType},
caption=rr:datatype to define the literal's datatype]
<#Website>      rml:reference "website" ;
§\label{rr:termType}§                rr:termType rr:IRI .
\end{lstlisting}

IRIs and constant values are generated
in the same way in RML and R2RML.
However, the \underline{blank nodes are generated differently}
in R2RML and RML:

\paragraph{R2RML} 
A \textit{Term Map} consists of \underline{one}
\textit{column} reference, 
\textit{template} or
\textit{constant}.

\paragraph{RML}
A \textit{Term Map} consists of
\underline{zero or one}
\textit{reference},
\textit{template} or
\textit{constant}.
This way,
blank nodes can be generated
relying on randomly generated IRIs.

\begin{lstlisting}[label={lst:rml_blankNode},
caption=Blank Nodes generated randomly without reference to the data source]
§\label{r2rml:bn}§<#RDFtermBlankNode>       rr:termType rr:BlankNode .

\end{lstlisting}

In \cref{tab:termTypeCombinations}, there is 
a list of all possible combinations of 
\textit{Term Maps}, their values and types,
as well as what type of RDF term is generated
in each case.

\begin{table}[ht]
	\centering
	\caption{All combinations of Term Maps and Term Types and the RDF terms they generate. The default term type for each term map is in parenthesis (optional to be specified).
	The most frequent used value and RDF type
	for each Term Map is in bold.
	[R2]RML is indicated when the statement is valid for both R2RML and RML.}
	\label{tab:termTypeCombinations}
	\begin{tabular}{c|c|c|l|l}
	\textbf{Term Map}
	& \textbf{language}
	& \textbf{value}
	& \textbf{Term Type}
	& \textbf{RDF Type} \\ \thickhline
	
	\multirow{7}{*}{\makecell[c]{Subject Map\\\texttt{rr:SubjectMap}}}
	& \multirow{4}{*}{[R2]RML}
	& \multirow{2}{*}{\texttt{\textbf{rr:template}}}
	& (\texttt{rr:IRI})
	& \textbf{IRI} \\ 
	 
	&
	&
	& \texttt{rr:BlankNode}
	& BlankNode \\ \cline{3-5}

    
    & 
	& \texttt{rr:constant}
	& (\texttt{rr:IRI})
	& IRI \\ \cline{3-5}
	
	
	& 
	& \texttt{rr:column}
	& \texttt{rr:IRI}
	& IRI \\ \cline{2-5}
	
	
	& \multirow{3}{*}{RML}
	& \multirow{2}{*}{\texttt{rml:reference}}
	& \texttt{rr:IRI}
	& IRI \\ 
	
	
	& 
	& 
	& \texttt{rr:BlankNode}
	& BlankNode \\ \cline{3-5}
	
	
	& 
	& --
	& \texttt{rr:BlankNode}
	& BlankNode \\ \thickhline
	
    \multirow{4}{*}{\makecell[c]{Predicate Map\\\texttt{rr:PredicateMap}}} 
    & \multirow{3}{*}{[R2]RML}
	& \texttt{rr:template}
	& (\texttt{rr:IRI})
	& IRI \\ \cline{3-5}

    
    & 
	& \texttt{\textbf{rr:constant}}
	& (\texttt{rr:IRI})
	& \textbf{IRI} \\ \cline{3-5}

	 
	& 
	& \texttt{rr:column}
	& \texttt{rr:IRI}
	& IRI \\ \cline{2-5}
	
	 
	& RML
	& \texttt{rml:reference}
	& \texttt{rr:IRI}
	& IRI \\ \thickhline
	
	\multirow{9}{*}{\makecell[c]{Object Map\\\texttt{rr:ObjectMap}}}
	& \multirow{6}{*}{[R2]RML}
	& \multirow{2}{*}{\texttt{rr:template}}
	& \textbf{IRI}
	& IRI \\ 
	
	 
	&
	& 
	& \texttt{rr:Literal}
	& Literal \\ \cline{3-5}
	
    
    & 
	& \multirow{2}{*}{\texttt{rr:constant}}
	& (\texttt{rr:IRI})
	& IRI \\ 

    
    & 
	& 
	& \texttt{rr:Literal}
	& IRI \\ \cline{3-5}
	 
    
    & 
	& \multirow{2}{*}{\texttt{\textbf{rr:column}}}
	& (\texttt{rr:Literal})
	& \textbf{Literal} \\ 
	
    
    & 
	& 
	& \texttt{rr:IRI}
	& IRI \\ \cline{2-5}

	
	& \multirow{3}{*}{RML}
	& \multirow{3}{*}{\texttt{\textbf{rml:reference}}}
	& \texttt{rr:IRI}
	& IRI \\ 


	&
	&
	& \texttt{rr:BlankNode}
	& BlankNode \\ 


	&
	&
	& \texttt{\textbf{rr:Literal}}
	& \textbf{Literal} \\ \thickhline
	
	\multirow{2}{*}{\makecell[c]{Referencing Object Map\\\texttt{rr:RefObjectMap}}}
	& \multirow{2}{*}{[R2]RML}
	& \multirow{2}{*}{\texttt{\textbf{rr:parentTriplesMap}}}
	& \textbf{IRI}
	& IRI \\ 
	
	&
	&
	& BlankNode
	& BlankNode \\ \hline
	
	\multirow{3}{*}{\makecell[c]{Language Map\\\texttt{rml:LanguageMap}}}
	& \multirow{3}{*}{RML}
	& \texttt{rr:template}
	& \texttt{rr:Literal}
	& n/a\\ \cline{3-5}
	
	& 
	& \texttt{rr:constant}
	& \texttt{rr:Literal}
	& n/a\\ \cline{3-5}
	
	& 
	& \texttt{rml:reference}
	& (\texttt{rr:Literal})
	& n/a\\ 

	\end{tabular}
\end{table}

\subsubsection{Data transformation}
The data often is not used
as they appear in the original data source,
but are transformed to generate the desired RDF term.
R2RML and RML follow different approaches
for transforing the data values.

\paragraph{R2RML}
A \underline{pre-processing} approach is considered.
If data is desired to be transformed,
the transformation occurs with
an SQL query or a view.

\paragraph{RML}
While pre-processing is possible,
not all data formats have
as powerful query languages as SQL is for relational databases.
If pre-processing occurs,
the data transformation is not declaratively defined.
Therefore, \underline{inline processing} approaches were proposed,
such as FnO~\cite{fno,demeester_eswc_2017}
and FunUL~\cite{ademar2016FunUL}.

\subsection{RDF triples generation}
\label{subsec:tripleGeneration}

In this subsection,
we explain how RDF terms are combined
to define RDF triples.
A subject, predicate, object and optionally a named graph
are Term Maps that are combined
to generate an RDF triple or quad respectively.
R2RML and RML follow the same way for generating RDF triples.

\subsubsection{Subject}
\label{subsec:subject}

The \textit{Subject Map} (\texttt{rr:SubjectMap})
defines how RDF terms are generated
that are unique identifiers (IRIs~\cite{iri})
or blank nodes.
These RDF terms constitute the subject of all RDF triples
generated from the \textit{Triples Map}.

\begin{lstlisting}[label={lst:r2rml_subjectMap},
caption=The Subject Map of a Triples Map]
<#PoleVaulters>     rr:subjectMap     <#Person_SM> .
§\label{r2rml:sm}§<#Person_SM>.       rr:template       "http:://ex.com/person/{name}"
\end{lstlisting}

\subsubsection{Predicate-Object}
\label{subsec:predicateObject}

A \textit{Predicate-Object Map} (\texttt{rr:PredicateObjectMap})
defines the pairs of predicates and objects
that characterise the \textit{subject}.
It consists of 
one or more mapping rules to define
how the \textit{predicate} (\texttt{rr:PredicateMap})
is generated, and 
one or more mapping rules to define
how the \textit{object} (\texttt{rr:ObjectMap})
or \textit{Referencing Object Maps} (\texttt{rr:ReferencingObjectMap})
is generated.

\begin{lstlisting}[label={lst:r2rml_pom},
caption=A Triples Map consists of zero or more Predicate Object Maps]
§\label{r2rml_triples_city:pom}§<#PoleVaulters>  rr:predicateObjectMap <#Mark_POM> ;
§\label{r2rml_triples_country:pom}§                    rr:predicateObjectMap <#Nationality_POM>. 

\end{lstlisting}

\subsubsection{Predicate}
A \textit{Predicate Map} (\texttt{rr:PredicateMap},
\cref{lst:r2rml_POM}, \cref{r2rml:pm})
is a \textit{Term Map} (\texttt{rr:TermMap}) 
that defines how the triple’s predicate is generated. 

\begin{lstlisting}[label={lst:r2rml_POM},
caption=A Predicate Object Map consists of one or more Predicate Maps\ and one or more Object Maps or Referencing Object Maps]
# Predicate Object Map with Object Map
§\label{r2rml:pm}§<#Mark_POM>       rr:predicate          ex:score ; 
§\label{r2rml:om}§                  rr:objectMap          [ rr:column "Mark"] .

# Predicate Object Map with Referencing Object Map
<#Nationality_POM>  rr:predicateMap       <#Country_PM> ; 
§\label{r2rml:rom}§                    rr:objectMap          <#Country_ROM> . 

\end{lstlisting}

\subsubsection{Object}
An \textit{Object Map} (\texttt{rr:ObjectMap}, \cref{lst:r2rml_POM}, \cref{r2rml:om})
or a \textit{Referencing Object Map} (\texttt{rr:ReferencingObjectMap}, \cref{lst:r2rml_POM}, \cref{r2rml:rom}) defines
how the triple’s object is generated.
An \textit{Object Map} is a \textit{Term Map}
that defines
how a resource (IRI or blank node) or a literal will be generated. 

\subsubsection{Referencing Object}

A \textit{Referencing Object Map} defines
how the object is generated based on the \textit{Subject Map} of another \textit{Triples Map}.
If the \textit{Triples Maps} refer to different \textit{Logical Tables},
a join between the \textit{Logical Tables} is required.
The \textit{join condition} (\texttt{rr:joinCondition})
performs the join exactly as a join is executed in SQL.
The join condition consists of a reference to a column name
that exists in the \textit{Logical Table} of the \textit{Triples Map}
that contains the \textit{Referencing Object Map} (\texttt{rr:child}) and
a reference to a column name
that exists in the \textit{Logical Table}
of the \textit{Referencing Object Map}’s \textit{Parent Triples Map} (\texttt{rr:parent}). 

\begin{lstlisting}[label={lst:r2rml_ROM},
caption=A Referencing Object Map generates an object based on the Subject Map of another Triples Map]
# Referencing Object Map
§\label{r2rml_term:ptm}§<#Country_ROM>   rr:parentTriplesMap <#Country_TM> ;
§\label{r2rml:jc}§                 rr:join  [ 
§\label{r2rml:jc_c}§                             rr:cild "nationality" ;
§\label{r2rml:jc_p}§                             rr:parent "country"] .
\end{lstlisting}

In both R2RML and RML,
the entity's RDF type and the RDF triple's named graph
may be defined on \textit{Subject Map}
or \textit{Predicate Object Map} level.

\subsubsection{RDF type}
In both R2RML and RML,
the RDF type of an element can be specified with two ways:
(i) specifying its class (\texttt{rr:class})
in the \textit{Subject Map} (\cref{lst:r2rml_class}, \cref{r2rml:class}),
or (ii) defining a \textit{Predicate Object Map} (\cref{lst:r2rml_POMclass})
whose predicate is \texttt{rdf:type} (\cref{r2rml:pm_class}) and
its object the desired class (\cref{r2rml:om_class}).

\begin{lstlisting}[label={lst:r2rml_class},
caption=The RDF type of an entity defined in the Subject Map]
<#Person_SM>  rr:template  "http:://ex.com/person/{name}" ;
§\label{r2rml:class}§              rr:class     foaf:Person .
\end{lstlisting}

\begin{lstlisting}[label={lst:r2rml_POMclass},
caption=The RDF type of an entity defined in the Predicate Object Map]
§\label{r2rml:pm_class}§<#Mark_POM>   rr:predicate. rdf:type ; 
§\label{r2rml:om_class}§              rr:object     foaf:Person .
\end{lstlisting}

\subsubsection{Named Graph} 
\label{subsec:namedGraph}
Each RDF triple is placed into
one or more graphs,
the unnamed default graph or an IRI-named named graph.
A \textit{Subject Map} or \textit{Predicate-Object Map}
may have one or more associated \textit{Graph Maps}. 

\begin{lstlisting}[label={lst:r2rml_graph},
caption=The named graph of the RDF triple may be defined in either Subject Map or Predicate Object Map]
<#Person_SM>. rr:template. "http:://ex.com/person/{name}_{surname}" ;
§\label{r2rml:graph}§              rr:graph  ex:PersonGraph .

<#Mark_POM>  rr:predicate          ex:score ; 
             rr:objectMap          [ rr:column "Mark"] ;
§\label{r2rml:gm}§             rr:graphMap           [ rr:constant <http://example.com/graph/students> ] .
\end{lstlisting}



\section{Rules Validation and Inconsistencies Resolution}
\label{sec:Validation}

Either R2RML or RML is considered
to define the rules 
for generating the desired knowledge graph,
the rules that define
how the knowledge graph is generated,
should be validated
before they are used to generate the knowledge graph.
This way, we prevent the same violations to appear repeatedly,
and inconsistencies are resolved in due time.
In \cref{subsec:validation},
we explain how the rules
can be validated and,
in \cref{subsec:resolution}
how inconsistencies can be resolved.

\subsection{Validation}
\label{subsec:validation}

Rules in both R2RML and RML constitute a knowledge graph,
because they have a native RDF representation.
Thus, the same set of schema validation patterns,
normally applied on the generated knowledge graphs,
is also applicable on the knowledge graph of the rules
that state how the knowledge graph is generated.
Therefore,
instead of validating the generated RDF triples,
we validate the \textit{Triples} and \textit{Term Maps}
that define how the RDF triples should be generated.

\subsubsection{RDF triples validation}
In the case of RDF triples validation,
the RDF triples are considered.
The RDF triple’s predicate (\texttt{foaf:familyName})
is validated against its subject's class
for domain violations
(\texttt{foaf:Person} in \cref{lst:correct_triples} and \texttt{foaf:Document} in \cref{lst:incorrect_triples})
and object (\texttt{"Morris"@en-us} in \cref{lst:correct_triples}
and \texttt{"Morris"}\textasciicircum\textasciicircum\texttt{xsd:integer} in \cref{lst:incorrect_triples})
for range violations.

\begin{lstlisting}[label={lst:correct_triples},
caption=RDF triples without violations.]
ex:Anzhelika_Sidorova a foaf:Person ; foaf:familyName "Sidorova"@en-us .
ex:Sandy_Morris       a foaf:Person ; foaf:familyName "Morris"@en-us .
ex:Katerina_Stefanidi a foaf:Person ; foaf:familyName "Stefanidi"@en-us .
ex:Holly_Bradshaw     a foaf:Person ; foaf:familyName "Bradshaw"@en-us .
\end{lstlisting}

\begin{lstlisting}[label={lst:incorrect_triples},
caption=RDF triples with violations.]
ex:Anzhelika_Sidorova a *foaf:Document* ; foaf:familyName *"Sidorova"^^xsd:integer* .
ex:Sandy_Morris       a *foaf:Document* ; foaf:familyName *"Morris"^^xsd:integer* .
ex:Katerina_Stefanidi a *foaf:Document* ; foaf:familyName *"Stefanidi"^^xsd:integer* .
ex:Holly_Bradshaw     a *foaf:Document* ; foaf:familyName *"Bradshaw"^^xsd:integer* .
\end{lstlisting}

While the RDF triples in \cref{lst:correct_triples}
do not have any violations,
the RDF triples in \cref{lst:incorrect_triples}
do have violations.
The RDF triples in the latter case may be corrected,
but the root of the violation is not known,
unless its provenance is traced.
But even then,
very fine-grained provenance information
is required even on RDF Terms level
\cite{dimou_ldow_2016,DeMeester2017Detailed}
to trace rules that generate violating RDF triples.

\subsubsection{rules validation}
In case of rules validation,
the predicate (\texttt{foaf:familyName})
is extracted from the \textit{Predicate Map}
(\cref{lst:correct_rules}, \cref{r2rml:correct_predicate}
in the former case and
\cref{lst:incorrect_rules}, \cref{r2rml:incorrect_predicate}
in the latter case)
and is compared to the ones derived from the corresponding
\textit{Subject Map}
(\cref{lst:correct_rules}, \cref{r2rml:correct_subject}
in the former case and
\cref{lst:incorrect_rules}, \cref{r2rml:incorrect_subject}
in the latter case)
and \textit{Object Map}
(\cref{lst:correct_rules}, \cref{r2rml:correct_object}
in the former case and
\cref{lst:incorrect_rules}, \cref{r2rml:incorrect_object}
in the latter case).
 are identified
The properties and classes namespaces
are used to retrieve the schemas and
generate the test cases.

\begin{lstlisting}[label={lst:correct_rules},
caption=Correct defined rules generate RDF triples without violations.]
<#PoleVaulters>     rr:subjectMap         <#Person_SM> ;
                    rr:predicateObjectMap <#Name_POM> .

<#Person_SM>.       rr:template       "http:://ex.com/person/{name}_{surname}" ;
§\label{r2rml:correct_subject}§                    rr:class          foaf:Person .
                    
§\label{r2rml:correct_predicate}§<#Name_POM>         rr:predicateMap [ foaf:familyName ];
§\label{r2rml:correct_object}§                    rr:objectMap    [ rml:reference "surname" ; rr:language "en-us" ] .
                    
\end{lstlisting}

\begin{lstlisting}[label={lst:incorrect_rules},
caption=Incorrect defined rules generate violating RDF triples.]
<#PoleVaulters> rr:subjectMap         <#Person_SM> ;
                rr:predicateObjectMap <#Name_POM> .

<#Person_SM>.   rr:template       "http:://ex.com/person/{name}_{surname}" ;
§\label{r2rml:incorrect_subject}§                rr:class          *foaf:Document* .
                    
§\label{r2rml:incorrect_predicate}§<#Name_POM>     rr:predicateMap [ foaf:familyName ];
§\label{r2rml:incorrect_object}§                rr:objectMap    [ rml:reference "surname" ; *rr:dataType xsd:integer* ] .
                    
\end{lstlisting}

While the rules in \cref{lst:correct_rules}
do not seem to generate RDF triples with violations,
the rules in \cref{lst:incorrect_rules}
will certainly generate RDF triples with violations.
The rules in the latter case may be corrected,
and the violations will never appear again,
unless new are introduced after the correction.

The RDF triples validation 
requires \underline{performing the validation as many times}
\underline{as the number of the generated RDF triples}.
In our example,
four RDF triples are generated
(\cref{lst:correct_triples} and \cref{lst:incorrect_triples}),
and zero and eight violations are identified respectively.
If there were four million RDF triples,
each triple should have been validated,
and zero and eight million violations
would have been identified respectively.
If the RDF triples are corrected,
the next time the RDF triples are generated,
the same violations will appear.

To the contrary, the rules validation is \underline{independent
of the number of the} \underline{generated RDF triples}.
In our example,
a \textit{Predicate Map} is compared with
its associated \textit{Subject Map} and \textit{Predicate Map}
and two violations are identified.
If the violating rules are fixed,
every next time that the RDF triples are generated,
the RDF triples will not have violations.

Even though assessing the rules can cover many violations
related to vocabularies and ontologies used to annotate the data,
some schema-related violations
depend on how the rules are instantiated on the original data. 
Therefore, a uniform way of incrementally assessing the quality of a knowledge graph and rules should cover both the rules and the knowledge graph.

\subsection{Resolution}
\label{subsec:resolution}

Resglass~\cite{Heyvaert2019Rule} includes a ranking
to determine the order with which rules and vocabulary terms
should be inspected.
Ranking inconsistencies helps in reducing 
the effort required
during the resolution of inconsistencies
in both rules and vocabularies.
The concrete steps are the following:

\begin{enumerate}
    \item \textit{Rules inconsistency detection}
    Inconsistencies in RML rules are detected
    via a rule-based reasoning system~\cite{DeMeester2019RDF};
    \item \textit{RML rules clustering} 
    RML rules are clustered according to the Triples Map
    to which the rules, i.e., Term Maps, correspond;
    \item \textit{RML rules ranking} 
    A score for every rules cluster and vocabulary term
    is calculated by
    iterating over each cluster that represents an entity
    and every Terms Map in the cluster 
    to count inconsistencies in which a rule is involved;
    \item \textit{RML rules refinement}
    experts inspect the top rules clusters, 
    and apply the necessary refinements to the RML rules;
    \item \textit{Knowledge graph generation}
    The knowledge graph is generated
    by applying the semantic annotations 
    to existing data sources via the RML rules; 
    \item \textit{Knowledge graph inconsistency detection}
    The knowledge graph is validated to determine inconsistencies.
    \item \textit{RML rules refinement}
    Inconsistencies detected in the previous steps
    might be resolved by refining the RML rules and vocabulary terms.
\end{enumerate}

\section{Conclusions}
In this paper,
we discuss in detail the differences in syntax
between the W3C recommended R2RML mapping language
and its most broadly used generalisation RML.
Both R2RML and RML may be used for generating
RDF from data in relational databases.
If a data owner only holds data in relational databases,
using R2RML should suffice.
However, 
if a dataowner holds data which is heterogeneous
and not in relational databases,
then RML is the only option.

While RML generalises R2RML,
it also inherits R2RML's limitations.
Thus, there are still aspects
that are not supported by none of the two languages,
such as RDF collections and containers
or nested term maps.
xR2RML~\cite{michel2016xr2rml}, an extension
over both R2RML and RML, is the only work
which proposes a solution towards this direction,
but more research is required.

As far as rules validation is concerned,
more research is required to further investigate
how rules can be validated and more importantly
how inconsistencies may be resolved.
Even though research is performed recently
aiming to increase the quality of knowledge graphs,
there is still room for improvement.
Approaches applied to the rules
that generate the knowledge graph
point exactly to the root causing the inconsistency
and prevent violations from being propagated.
They also require significantly less time and resources,
as opposed to time-consuming and performance-intensive
approaches applied to the generated knowledge graph.
Even though preliminary efforts
propose methodologies to resolve inconsistencies,
more research is still required to achieve
more automated approaches
that require less human intervention and 
contribute in generating higher quality knowledge graphs.

%
\bibliographystyle{abbrv}
\bibliography{bib.bib}

\begin{thebibliography}{10}

\bibitem{Acosta2013}
M.~Acosta, A.~Zaveri, E.~Simperl, D.~Kontokostas, S.~Auer, and J.~Lehmann.
\newblock {Crowdsourcing linked data quality assessment}.
\newblock In {\em The Semantic Web}, 2013.

\bibitem{sparql_spec}
C.~B. Aranda, O.~Corby, S.~Das, L.~Feigenbaum, P.~Gearon, B.~Glimm, S.~Harris,
  S.~Hawke, I.~Herman, N.~Humfrey, N.~Michaelis, C.~Ogbuji, M.~Perry,
  A.~Passant, A.~Polleres, E.~Prud'hommeaux, A.~Seaborne, and G.~T. Williams.
\newblock {SPARQL 1.1 Overview}.
\newblock Recommendation, World Wide Web Consortium (W3C), 2013.

\bibitem{Arenas2012}
M.~Arenas, A.~Bertails, E.~Prud'hommeaux, and J.~Sequeda.
\newblock {A Direct Mapping of Relational Data to RDF}.
\newblock Recommendation, World Wide Web Consortium (W3C), 2012.

\bibitem{Bizer2009c}
C.~Bizer and R.~Cyganiak.
\newblock {Quality-driven information filtering using the WIQA policy
  framework}.
\newblock {\em Web Semantics: Science, Services and Agents on the World Wide
  Web}, 7(1):1--10, 2009.

\bibitem{Calvanese_2011}
D.~Calvanese, G.~De~Giacomo, D.~Lembo, M.~Lenzerini, A.~Poggi,
  M.~Rodriguez-Muro, R.~Rosati, M.~Ruzzi, and D.~F. Savo.
\newblock The mastro system for ontology-based data access.
\newblock {\em Semantic Web}, 2011.

\bibitem{Catherine_2019_WeatherDP}
R.~Catherine, B.~Stephan, A.~G{\'e}raldine, and B.~Daniel.
\newblock Weather data publication on the lod using sosa / ssn ontology.
\newblock {\em Semantic Web}, 2019.

\bibitem{D2RQ}
R.~Cyganiak, C.~Bizer, J.~Garbers, O.~Maresch, and C.~Becker.
\newblock The {D2RQ} {M}apping {L}anguage.
\newblock Technical report, FU Berlin, DERI, UCB, JP Morgan Chase, AGFA, HP
  Labs, Johannes Kepler Universität Linz, 2012.

\bibitem{RDF}
R.~Cyganiak, D.~Wood, and M.~Lanthaler.
\newblock {RDF 1.1 Concepts and Abstract Syntax}.
\newblock Recommendation, World Wide Web Consortium (W3C), 2014.

\bibitem{R2RML}
S.~Das, S.~Sundara, and R.~Cyganiak.
\newblock {R2RML: RDB to RDF Mapping Language}.
\newblock Working group recommendation, World Wide Web Consortium (W3C), 2012.

\bibitem{fno}
B.~De~Meester and A.~Dimou.
\newblock {T}he {F}unction {O}ntology.
\newblock {U}nofficial {D}raft, Ghent University -- imec -- IDLab, 2016.

\bibitem{DeMeester2017Detailed}
B.~De~Meester, A.~Dimou, R.~Verborgh, and E.~Mannens.
\newblock Detailed provenance capture of data processing.
\newblock In D.~Garijo, W.~R. van Hage, T.~Kauppinen, T.~Kuhn, and J.~Zhao,
  editors, {\em Proceedings of the First Workshop on Enabling Open Semantic
  Science (SemSci)}, volume 1931 of {\em CEUR Workshop Proceedings}, 2017.

\bibitem{DeMeester2019RDF}
B.~De~Meester, P.~Heyvaert, D.~Arndt, A.~Dimou, and R.~Verborgh.
\newblock {RDF Graph Validation Using Rule-Based Reasoning}.
\newblock {\em Semantic Web Journal}, 2020.
\newblock Accepted.

\bibitem{DeMeester2019Mapping}
B.~De~Meester, P.~Heyvaert, R.~Verborgh, and A.~Dimou.
\newblock {Mapping languages analysis of comparative characteristics}.
\newblock In D.~Chaves-Fraga, P.~Heyvaert, F.~Priyatna, J.~Sequeda, A.~Dimou,
  H.~Jabeen, D.~Graux, G.~Sejdiu, M.~Saleem, and J.~Lehmann, editors, {\em
  Joint Proceedings of the 1\textsuperscript{st} International Workshop on
  Knowledge Graph Building and 1\textsuperscript{st} International Workshop on
  Large Scale RDF Analytics co-located with 16\textsuperscript{th} Extended
  Semantic Web Conference (ESWC)}, volume 2489 of {\em CEUR Workshop
  Proceedings}, pages 37--45, Portoro\u{z}, Slovenia, 2019.

\bibitem{demeester_eswc_2017}
B.~De~Meester, W.~Maroy, A.~Dimou, R.~Verborgh, and E.~Mannens.
\newblock Declarative data transformations for {Linked Data} generation: the
  case of {DBpedia}.
\newblock In {\em Proceedings of the 14\textsuperscript{th} ESWC}, volume 10250
  of {\em LNCS}. Springer, 2017.

\bibitem{dimou_ldow_2016}
A.~Dimou, T.~De~Nies, R.~Verborgh, E.~Mannens, and R.~Van~de Walle.
\newblock Automated metadata generation for {L}inked {D}ata generation and
  publishing workflows.
\newblock In S.~Auer, T.~Berners-Lee, C.~Bizer, and T.~Heath, editors, {\em
  Proceedings of the Workshop on Linked Data on the Web co-located with
  25\textsuperscript{th} International World Wide Web Conference (WWW2016)},
  CEUR Workshop Proceedings, 2016.

\bibitem{dimou2015assessing}
A.~Dimou, D.~Kontokostas, M.~Freudenberg, R.~Verborgh, J.~Lehmann, E.~Mannens,
  S.~Hellmann, and R.~Van~de Walle.
\newblock {Assessing and Refining Mappings to RDF to Improve Dataset Quality}.
\newblock pages 133--149, 2015.

\bibitem{dimou_ldow_2014}
A.~Dimou, M.~Vander~Sande, P.~Colpaert, R.~Verborgh, E.~Mannens, and R.~Van~de
  Walle.
\newblock {RML:} {A} {G}eneric {L}anguage for {I}ntegrated {RDF} {M}appings of
  {H}eterogeneous {D}ata.
\newblock In {\em Proceedings of the 7\textsuperscript{th} Workshop on Linked
  Data on the Web}, volume 1184 of {\em CEUR Workshop Proceedings}. CEUR, 2014.

\bibitem{Dimou_2015_LogicalSource}
A.~Dimou, R.~Verborgh, M.~V. Sande, E.~Mannens, and R.~Van~de Walle.
\newblock Machine-interpretable dataset and service descriptions for
  heterogeneous data access and retrieval.
\newblock In {\em Proceedings of the 11th International Conference on Semantic
  Systems}, SEMANTICS '15, pages 145--152. ACM, 2015.

\bibitem{iri}
M.~Duerst and M.~Suignard.
\newblock {Internationalized Resource Identifiers (IRIs)}.
\newblock Standard track, IETF, 2005.

\bibitem{farid2016clams}
M.~Farid, A.~Roatis, I.~F. Ilyas, H.-F. Hoffmann, and X.~Chu.
\newblock {CLAMS: Bringing Quality to Data Lakes}.
\newblock In F.~\"{O}zcan and G.~Koutrika, editors, {\em Proceedings of the
  2016 International Conference on Management of Data (SIGMOD)}, pages
  2089--2092, New York, United States, 2016. ACM.

\bibitem{Haase2019Hybrid}
P.~Haase.
\newblock {Hybrid Enterprise Knowledge Graphs}.
\newblock Technical report, metaphacts GmbH, 2019.

\bibitem{Heyvaert2018Declarative}
P.~Heyvaert, B.~De~Meester, A.~Dimou, and R.~Verborgh.
\newblock {Declarative Rules for Linked Data Generation at your Fingertips!}
\newblock 2018.

\bibitem{Heyvaert2019Rule}
P.~Heyvaert, A.~Dimou, B.~De~Meester, and R.~Verborgh.
\newblock {Rule-driven inconsistency resolution for knowledge graph generation
  rules}.
\newblock {\em Semantic Web Journal}, 10(6):1071--1086, 2019.

\bibitem{Hogan:2012:ESL:2263498.2264570}
A.~Hogan, J.~Umbrich, A.~Harth, R.~Cyganiak, A.~Polleres, and S.~Decker.
\newblock {An Empirical Survey of Linked Data Conformance}.
\newblock {\em Journal of Web Semantics}, 2012.

\bibitem{ademar2016FunUL}
A.~C. Junior, C.~Debruyne, R.~Brennan, and D.~O’Sullivan.
\newblock Funul: A method to incorporate functions into uplift mapping
  languages.
\newblock In {\em Proceedings of the 18th International Conference on
  Information Integration and Web-Based Applications and Services}, page
  267–275, New York, USA, 2016. ACM.

\bibitem{Kontokostas2014}
D.~Kontokostas, P.~Westphal, S.~Auer, S.~Hellmann, J.~Lehmann, R.~Cornelissen,
  and A.~Zaveri.
\newblock {Test-driven evaluation of linked data quality}.
\newblock In C.-W. Chung, editor, {\em Proceedings of the
  23\textsuperscript{rd} international conference on World Wide Web}, pages
  747--757, New York, United States, 2014. ACM.

\bibitem{Lange_2011_Krextor}
C.~Lange.
\newblock {Krextor - An Extensible Framework for Contributing Content Math to
  the Web of Data}.
\newblock In J.~H. Davenport, W.~M. Farmer, J.~Urban, and F.~Rabe, editors,
  {\em Intelligent Computer Mathematics}, pages 304--306. Springer, 2011.

\bibitem{Langegger_2009_XLWrap}
A.~Langegger and W.~W{\"o}{\ss}.
\newblock {XLWrap -- Querying and Integrating Arbitrary Spreadsheets with
  SPARQL}.
\newblock In {\em The Semantic Web - ISWC 2009}. Springer, 2009.

\bibitem{lefrancois_eswc_2017}
M.~Lefran\c{c}ois, A.~Zimmermann, and N.~Bakerally.
\newblock A {SPARQL} extension for generating {RDF} from heterogeneous formats.
\newblock In {\em The Semantic Web 14\textsuperscript{th} International
  Conference, ESWC 2017, Proceedings}, Portoroz, Slovenia, 2017. Springer.

\bibitem{Lehmann2015DBpediaA}
J.~Lehmann, R.~Isele, M.~Jakob, A.~Jentzsch, D.~Kontokostas, P.~N. Mendes,
  S.~Hellmann, M.~Morsey, P.~van Kleef, S.~Auer, and C.~Bizer.
\newblock {DBpedia -- A large-scale, multilingual knowledge base extracted from
  Wikipedia}.
\newblock {\em Semantic Web}, 6(2), 2015.

\bibitem{Makela_2012}
E.~Makela, E.~Hyv\"{o}nen, and T.~Ruotsalo.
\newblock How to deal with massively heterogeneous cultural heritage data:
  Lessons learned in culturesampo.
\newblock {\em Semantic Web}, 2012.

\bibitem{michel2015translation}
F.~Michel, L.~Djimenou, C.~Faron-Zucker, and J.~Montagnat.
\newblock {Translation of Heterogeneous Databases into RDF, and Application to
  the Construction of a SKOS Taxonomical Reference}.
\newblock In {\em International Conference on Web Information Systems and
  Technologies}, pages 275--296. Springer, 2015.

\bibitem{michel2016xr2rml}
F.~Michel, L.~Djimenou, C.~Faron-Zucker, and J.~Montagnat.
\newblock {xR2RML: Relational and Non-Relational Databases to RDF Mapping
  Language}.
\newblock Rapport de recherche, Laboratoire d'Informatique, Signaux et
  Systèmes de Sophia-Antipolis (I3S), 2017.

\bibitem{Patel-Schneider2014Using}
P.~F. Patel-Schneider.
\newblock {Using Description Logics for RDF Constraint Checking and
  Closed-World Recognition}.
\newblock In B.~Bonet and S.~Koenig, editors, {\em Proceedings of the
  29\textsuperscript{th} AAAI Conference on Artificial Intelligence}, pages
  247--253. AAAI Press, 2015.

\bibitem{paulheim_eswc_2017}
H.~Paulheim.
\newblock Data-driven {J}oint {D}ebugging of the {DB}pedia {M}appings and
  {O}ntology.
\newblock 2017.

\bibitem{scharffe:hal-00768424}
F.~Scharffe, G.~Atemezing, R.~Troncy, F.~Gandon, S.~Villata, B.~Bucher,
  F.~Hamdi, L.~Bihanic, G.~K{\'e}p{\'e}klian, F.~Cotton, J.~Euzenat, Z.~Fan,
  P.-Y. Vandenbussche, and B.~Vatant.
\newblock {Enabling Linked Data publication with the Datalift platform}.
\newblock In {\em AAAI Workshop on Semantic Cities}, Toronto, ON, Canada, 2012.

\end{thebibliography}

\end{document}